\documentclass[letterpaper, 10 pt, conference]{ieeeconf}  
\makeatletter
\let\NAT@parse\undefined
\makeatother

\IEEEoverridecommandlockouts                              

\let\labelindent\relax

\usepackage[utf8]{inputenc} 
\usepackage{amsmath,amssymb,amsfonts}
\usepackage{algorithm}
\usepackage{algpseudocode}
\usepackage{graphicx}
\usepackage{tabularx}
\usepackage{booktabs}
\usepackage{makecell}
\usepackage{textcomp}
\usepackage{xcolor}
\usepackage{lipsum}
\usepackage{enumitem}
\usepackage[numbers]{natbib}

\newcolumntype{Y}{>{\centering\arraybackslash}X}
\def\BibTeX{{\rm B\kern-.05em{\sc i\kern-.025em b}\kern-.08em
    T\kern-.1667em\lower.7ex\hbox{E}\kern-.125emX}}

\title{\LARGE \bf Training Effective Neural Sentence Encoders from Automatically Mined Paraphrases}
\author{Sławomir Dadas \\
\textit{National Information Processing Institute}, Warsaw, Poland}

\begin{document}

\maketitle
\thispagestyle{empty}
\pagestyle{empty}

\begin{abstract}
Sentence embeddings are commonly used in text clustering and semantic retrieval tasks. State-of-the-art sentence representation methods are based on artificial neural networks fine-tuned on large collections of manually labeled sentence pairs. Sufficient amount of annotated data is available for high-resource languages such as English or Chinese. In less popular languages, multilingual models have to be used, which offer lower performance. In this publication, we address this problem by proposing a method for training effective language-specific sentence encoders without manually labeled data. Our approach is to automatically construct a dataset of paraphrase pairs from sentence-aligned bilingual text corpora. We then use the collected data to fine-tune a Transformer language model with an additional recurrent pooling layer. Our sentence encoder can be trained in less than a day on a single graphics card, achieving high performance on a diverse set of sentence-level tasks. We evaluate our method on eight linguistic tasks in Polish, comparing it with the best available multilingual sentence encoders.
\end{abstract}

\section{Introduction}
Using artificial neural networks for generating dense vector representations of text has become a common practice in natural language processing (NLP). While most research is focused on word and subword representations, some applications require encoding larger chunks of text, such as sentences or paragraphs. Neural sentence encoders are used in semantic search, question answering, document clustering, dataset augmentation, plagiarism detection, and other tasks which involve measuring semantic similarity between sentences. Typically, these types of models are utilized for transforming text fragments to their corresponding dense representations, which are then processed independently by a vector search engine or other information retrieval system. This allows billions of vectors to be compared and searched efficiently. The quality of these representations, therefore, has a significant impact on the performance of the whole system. 

In recent years, a number of methods for encoding sentences have been introduced. State-of-the-art models, producing high-quality vector representations, are usually trained in a supervised way. Training semantically meaningful representations, suitable for search and retrieval problems, requires specific types of labeled datasets such as natural language inference (NLI) or paraphrase pairs. Large datasets of manually labeled sentence pairs exist only for high-resource languages. For English, SNLI \citep{bowman2015large} and MultiNLI \citep{williams2018broad} datasets are available, among others, each with several hundred thousand records. For other languages, there are either no NLI datasets, or the size of existing datasets is insufficient for training high-quality neural sentence encoders. Therefore, for low-resource languages, multilingual or unsupervised sentence encoders need to be used, which offer lower performance than the best methods trained for English.

This publication addresses this problem by proposing a method for training effective language-specific sentence encoders without manually labeled data. In the first step, we perform automatic extraction of paraphrase pairs in the target language using a large corpus of parallel sentences from the OPUS project \citep{tiedemann2012parallel}. Next, we fine-tune a siamese network composed of two Transformer-based \citep{vaswani2017attention} language models on the collected data to discriminate paraphrases from non-paraphrases. For fine-tuning, we employ an additional LSTM \citep{hochreiter1997long} (Long Short-term Memory) layer as a pooling operation of which the last hidden state is used as sentence representation. We show that it is possible to train a neural sentence encoder for a lower resource language on a single GPU in under 24 hours, achieving better performance than state-of-the-art multilingual models trained with significantly more data and compute power.

\section{Contributions}

We make the following contributions in this work:
\begin{enumerate}[wide,labelwidth=0pt,labelindent=0pt]
\item We propose a framework for training neural sentence encoders which does not require manually labeled data. Our method involves automatic extraction of paraphrase pairs for a target language utilizing sentence-aligned cross-lingual corpora. The resulting dataset is then used for fine-tuning Transformer language model to produce high-quality dense representations of sentences.
\item Our architecture generates sentence vectors from an additional LSTM pooling layer. Other popular Transformer-based sentence encoders use simple non-parametric pooling operations such as mean or max pooling, which restrict the dimension of the resulting vector. Using recurrent layer allows us to produce arbitrarily sized sentence representations.
\item We validate our approach on the Polish language. We conduct an evaluation on eight tasks, including the following problems: paraphrase identification, sentiment analysis, natural language inference, semantic relatedness, and topic classification. 
\item For the purposes of the evaluation, we publish the Polish Paraphrase Corpus (PPC). It is a new dataset consisting of 7000 manually labeled sentence pairs from different sources, each assigned to one of three categories: exact paraphrases, close paraphrases, non-paraphrases. Most of the examples have high semantic overlap, which makes the task challenging for classification models.
\item We make the source code for paraphrase mining, model fine-tuning, and evaluation publicly available\footnote{https://github.com/sdadas/polish-sentence-evaluation}. The code allows training sentence encoders and replicating our results for any language.
\end{enumerate}

\section{Related work}
Since the popularization of word embedding models such as Word2Vec \citep{mikolov2013efficient,mikolov2013distributed}, GloVe \citep{pennington2014glove}, or FastText \citep{bojanowski2017enriching}, there have been efforts to develop effective vector representations for larger fragments of text. Early approaches employed simple aggregation techniques of the individual word vectors, usually by computing arithmetic or weighted mean. More advanced methods based on static word representations have also been developed. \citet{arora2017a} introduced \emph{Smooth Inverse Frequency (SIF)} which included a weighted mean combined with principal component analysis (PCE), \citet{shen2018baseline} showed that concatenating mean and max pooled vectors improves the quality of sentence embeddings. Aggregation-based approaches offer intuitive and easy-to-use baselines, but the quality of the resulting representations is inferior to more recent models. 

Most of the modern methods are based on artificial neural networks trained on sentence-level optimization objective. Some of these models employ self-supervised learning and can be trained using only raw text corpora, while others are fully supervised and require labeled datasets for training. The model proposed by \citet{le2014distributed} was one of the first self-supervised approaches. It is a simple neural architecture for learning fixed-length paragraph vectors from word embedding models. The network was trained to predict the next word in the document from the representation of previously encoded words. Skip-Thought vectors \citep{kiros2015skipthought} is another notable example of self-supervised methods. It is an encoder-decoder network in which the input sentence is first encoded to a dense representation, and then the model is expected to reconstruct the previous and the next sentence from the same document. This method was followed by other similar architectures that improved on the original model \cite{gan2017learning,logeswaran2018an}. Recent neural sentence encoders typically do not involve optimizing the model from scratch, and they instead rely on pre-trained language models, the most popular of which are Transformer-based models. Latest self-supervised learning approaches propose fine-tuning a language model using denoising or contrastive objectives. TSDAE \citep{wang-etal-2021-tsdae-using} tries to reconstruct the original sentence from a damaged input, SimCSE \citep{gao-etal-2021-simcse} learns to identify two versions of the same sentence encoded with different dropout masks, Contrastive Tension \citep{carlsson2021semantic} trains two independent models on a noise-contrastive task. 

State-of-the-art sentence embedding models, achieving the highest performance on semantic retrieval tasks, are optimized using supervised learning. Some of the earlier popular models of this type include InferSent \citep{conneau2017supervised} and Universal Sentence Encoder \citep{cer2018universal}, both trained on the SNLI \citep{bowman2015large} corpus. Recently, several new Transformer-based approaches to learning sentence encoders have been developed as a part of the Sentence-Transformers library\footnote{https://www.sbert.net/}. Original method \citep{reimers2019sentence} is based on siamese neural network architecture composed of two Transformer models with shared parameters. Fine-tuning the network involves minimizing the distance between representations for similar sentences and maximizing for different sentences. The authors experimented with several loss functions, including cross-entropy loss, mean squared error loss on cosine similarity between vectors, triplet loss, multiple negatives ranking loss. The last one proved to produce the highest quality sentence embeddings. Pre-trained models are provided along with the library, the most recent of which were trained on a dataset of over one billion English sentence pairs.

The availability of pre-trained models and labeled datasets of sentence pairs is lower for languages other than English. Currently, the best option for these languages is to use multilingual sentence encoders, which offer reasonably good performance for semantic tasks. We can consider multilingual models as a separate group of methods since they are usually learned in a different way than the approaches described above. More specifically, these models are trained using big cross-lingual text corpora, exploiting semantic similarity between aligned sentence pairs in different languages. Since a large volume of cross-lingual data is available on the Internet, training an effective multilingual model is often computationally intensive and requires at least several hundred gigabytes of text. In recent years, a few pre-trained multilingual sentence encoders were published. \citet{10.1162/tacl_a_00288} released LASER, neural sentence encoder which can handle 93 languages. A multilingual version of Universal Sentence Encoder supporting 16 languages \citep{yang-etal-2020-multilingual} has also been made publicly available. LaBSE \citep{feng2020language} is another popular model, based on a multilingual BERT \citep{devlin-etal-2019-bert}, fine-tuned for semantic retrieval tasks in 112 languages. A different approach to training multilingual encoders has been shown in \citet{reimers-gurevych-2020-making}. The publication proposes a method for transferring knowledge from a pre-trained English sentence encoder (teacher) to a pre-trained multilingual language model (student) by minimizing the distance between English sentence vector and sentence vectors corresponding to translated sentences. This fine-tuning procedure requires less data than training multilingual encoders from scratch. Several pre-trained models created with this method have already been published in Sentence-Transformers library.

\begin{figure*}
  \centering
  \includegraphics[scale=0.70]{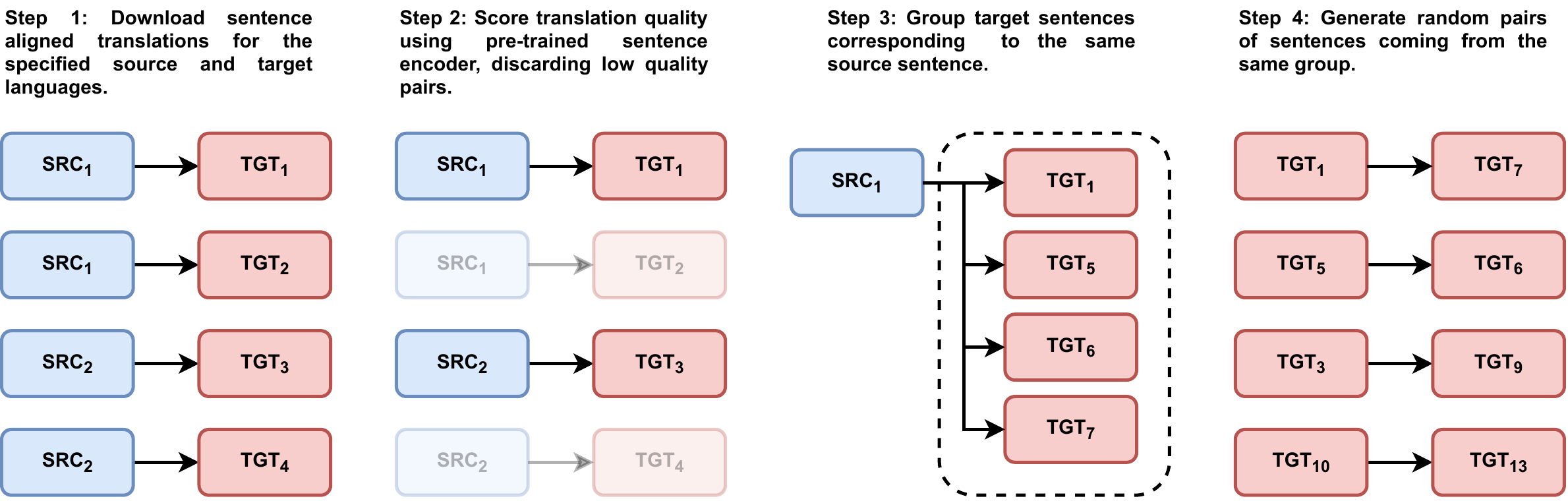}
  \caption{A diagram showing our procedure for paraphrase extraction from a bilingual corpus.}
  \label{fig:algorithm}
\end{figure*}

\section{Methodology}
In this section, we describe our approach to training Transformer-based sentence encoders from automatically mined paraphrases. First, we characterize our method of extracting paraphrase pairs from parallel corpora. Next, we present the architecture of the neural model employed in this study and describe the procedure of training the model on the collected data.

\subsection{Paraphrase extraction}
The main idea of our paraphrase extraction approach is to utilize sentence-aligned cross-lingual text corpora. One of the largest available collections of such corpora is the OPUS project \citep{tiedemann2012parallel}, which contains several dozen datasets covering almost all modern languages. In this study, two of them were used, which proved to work best with our method - OpenSubtitles \citep{lison-tiedemann-2016-opensubtitles2016} and CCMatrix \citep{schwenk-etal-2021-ccmatrix}. They both include multiple alternative translations for the same sentence, a characteristic that can be exploited for paraphrase mining.

A single run of our algorithm extracts paraphrased from a bilingual dataset. First, we select the source and target language. The target language is the one on which the model is trained, while the source language can be chosen arbitrarily. Usually, it is preferred to use English as the source language since most bilingual data is available for it. After downloading the sentence-aligned corpus, we perform data filtering step using a pre-trained multilingual sentence encoder. This step is necessary to improve the quality of resulting paraphrases because the original corpus is often noisy, containing translation errors and misaligned sentences. We compute the cosine similarity between a pair of sentences and discard pairs that are below a certain threshold. In our experiments, we used \emph{paraphrase-xlm-r-multilingual-v1} model from the Sentence-Transformers library to filter the data, and set a threshold of 0.7. In the next step, we group sentences in the target language corresponding to the same source sentences. We can then generate paraphrases from all groups containing at least two sentences. Within each group, we randomly select sentence pairs such that the resulting pairs contain at least one occurrence of each sentence. The dataset created in this way can then be used to train a sentence encoder. The procedure described above is shown in a graphical form in Figure \ref{fig:algorithm}.

\subsection{Neural sentence encoder}
Currently, a common approach to training sentence encoders is to use a pre-trained language model such as BERT \citep{devlin-etal-2019-bert} or RoBERTa \citep{liu2019roberta}. Such models already contain sufficient semantic knowledge, and they only need to be fine-tuned to efficiently encode representations of sentences for semantic retrieval tasks. In this study, we employ an approach similar to the one proposed by \citet{reimers2019sentence}. We create a siamese network composed of two Transformer models with tied weights, initialized with a pre-trained language model. This architecture is then trained on a dataset of sentence pairs. Each Transformer network independently produces a sentence vector, and the vectors are compared using a similarity function. The sentence representation is generated by the model from individual token embeddings using a pooling operation. \citet{reimers2019sentence} experimented with simple pooling strategies such as using the first (\emph{CLS}) token, computing the mean or max operation on all token vectors. These pooling methods are fast to compute but have a significant drawback - they restrict the size of the sentence embedding to be the same as the size of token vectors.

\begin{figure}
  \centering
  \includegraphics[scale=0.70]{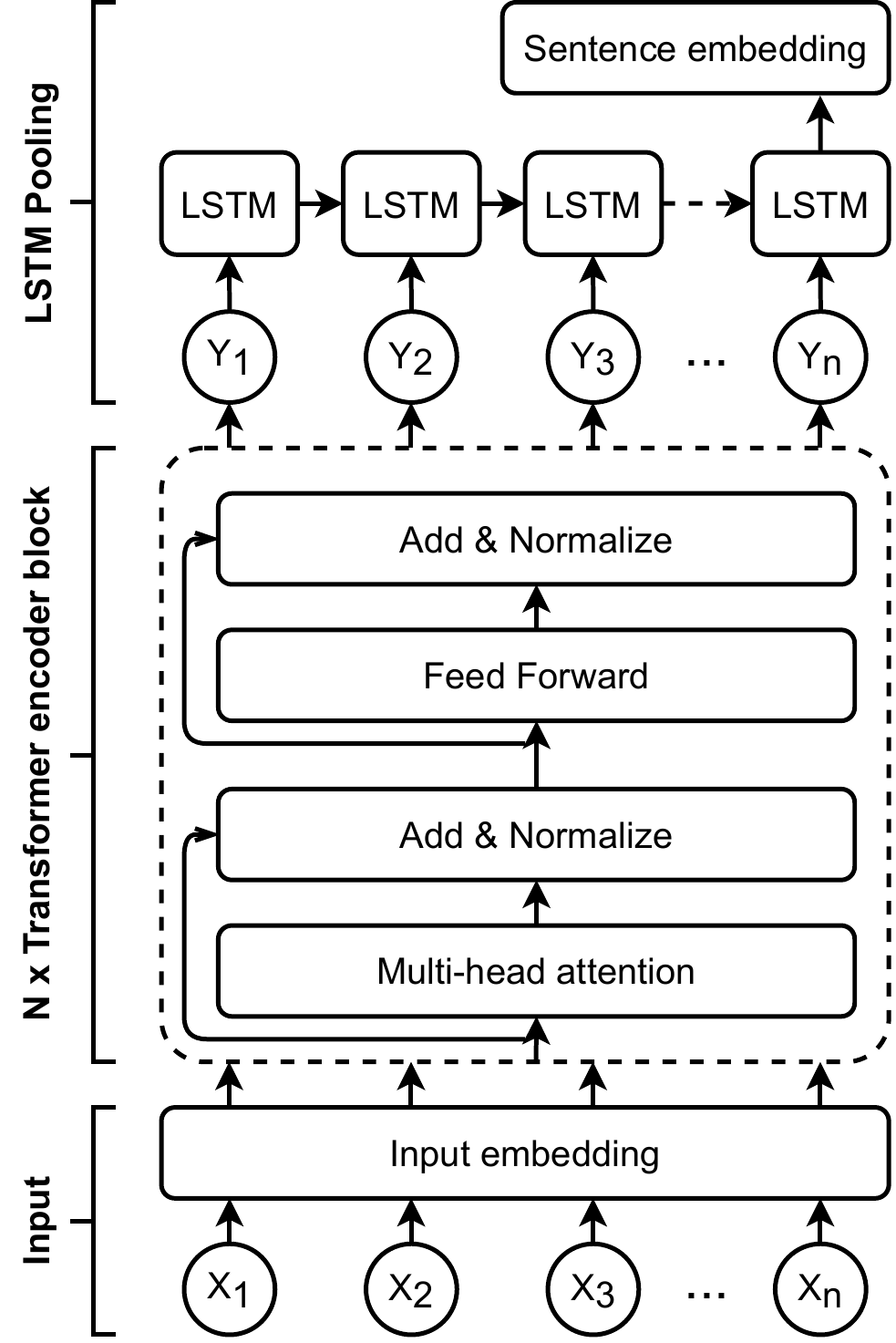}
  \caption{The architecture of our neural sentence encoder. The model is based on a standard Transformer architecture with an embedding layer and a number of self-attention blocks. The input of the model is a sequence of tokens $x_{1},...,x_{n}$, and the last of the encoder blocks outputs a vector representation of each token $y_{1},...,y_{n}$. The sentence representation is built from the individual token representations using an additional LSTM layer. The last hidden state of the recurrent layer is used as sentence embedding.}
  \label{fig:model}
\end{figure}

We believe that higher-dimensional sentence representations would preserve more semantic information, allowing higher performance on some tasks. Therefore, we propose a pooling operation based on an additional LSTM layer. The layer is placed after the last encoder block of the Transformer model. It takes a sequence of token vectors as input and returns a single vector representing the whole sentence. For the sentence embedding, we use the last hidden state of the LSTM cell. The architecture of the proposed sentence encoder is shown in Figure \ref{fig:model}. This approach allows an arbitrary number of dimensions to be set for the resulting sentence vector, since the size of the LSTM cell does not depend on the size of the input.

We fine-tune the model with mini-batch version of \emph{AdamW} \citep{loshchilov2018decoupled} algorithm. Every batch of size $K$ is composed of sentence pairs represented by their embeddings encoded by the model:
\begin{equation}
X=[(a_{1},b_{i}),(a_{2},b_{2}),...,(a_{K},b_{K})]
\end{equation}
Let us call the first sentence in each pair the \emph{anchor}. The second sentence in the same pair is its \emph{positive} sentence and the second sentences in all other pairs are its \emph{negative} sentences. During training, every \emph{anchor} $a_{i}$ is compared to all representations $b_{1},b_{2},...,b_{K}$. We expect the similarity of vectors to be high only for the positive pairs and low for all other pairs. Specifically, we use multiple negatives ranking loss function \citep{Henderson2017EfficientNL}:
\begin{equation}
J(X,\theta) = -\frac{1}{K}\sum_{i=1}^{K} \Big( sim(a_{i},b_{i}) - \log \sum_{j=1}^{K} e^{sim(a_{i},b_{j})} \Big)
\end{equation}
where X is the current mini-batch, $\theta$ denotes the model parameters, and $sim(a_{i},b_{j})$ is a similarity function comparing sentence representations $a_{i}$, $b_{j}$. In our case, the cosine similarity is used as the similarity function, defined as follows:
\begin{equation}
sim(x,y)=\frac{\sum_{i=1}^{n} x_{i} y_{i}}{\sqrt{\sum_{j=1}^{n} x_{j}^{2}} \sqrt{\sum_{j=1}^{n} y_{j}^{2}}}
\end{equation}
where both $x$ and $y$ are vectors, and $x_{i}$ denotes the $i$-th dimension of $x$.

\section{Experiments}
In this section, we demonstrate the results of our experiments. The method presented in this paper was evaluated on eight linguistic tasks in Polish and compared with other publicly available neural sentence encoders. In our experiments, we follow the evaluation approach from SentEval \citep{conneau-kiela-2018-senteval} toolkit. The evaluated sentence encoders are not fine-tuned for specific tasks, they are only used to generate sentence embeddings. For each task, a simple neural network with one hidden layer is trained, which takes these static sentence representations as input and outputs a class label or regression score. For classification tasks, we use accuracy as the evaluation metric. For semantic relatedness, Spearman's rank correlation coefficient is used.

For the purpose of our experiments, we created a new manually annotated dataset for paraphrase identification: Polish Paraphrase Corpus. First, we describe the corpus and the process of its development. Next, we present other datasets used in the evaluation. Then we present a description of the neural sentence encoders for Polish trained by us and other methods on which the evaluation was performed. The section concludes with a discussion of the results.

\subsection{Polish Paraphrase Corpus}
Polish Paraphrase Corpus contains 7000 manually labeled sentence pairs. The dataset was divided into training, validation and test splits. The training part includes 5000 examples, while the other parts contain 1000 examples each. The main purpose of creating such a dataset was to verify how machine learning models perform in the challenging problem of paraphrase identification, where most records contain semantically overlapping parts. Technically, this is a three-class classification task, where each record can be assigned to one of the following categories:
\begin{itemize}[wide,labelwidth=0pt,labelindent=0pt]
\item \emph{Exact paraphrases} - Sentence pairs that convey exactly the same information. We are interested only in the semantic meaning of the sentence, therefore this category also includes sentences that are semantically identical but, for example, have different emotional emphasis.
\item \emph{Close paraphrases} - Sentence pairs with similar semantic meaning. In this category we include all pairs which contain the same information, but in addition to it there may be other semantically non-overlapping parts. This category also contains context-dependent paraphrases - sentence pairs that may have the same meaning in some contexts but are different in others.
\item \emph{Non-paraphrases} - All other cases, including contradictory sentences and semantically unrelated sentences.
\end{itemize}

The corpus contains 2911, 1297, and 2792 examples for the above three categories, respectively. The process of annotating the dataset was preceded by an automated generation of candidate pairs, which were then manually labeled. We experimented with two popular techniques of generating possible paraphrases: backtranslation with a set of neural machine translation models and paraphrase mining using a pre-trained multilingual sentence encoder. The extracted sentence pairs are drawn from different data sources: Taboeba\footnote{
https://tatoeba.org/}, Polish news articles, Wikipedia and Polish version of SICK dataset \citep{dadas-etal-2020-evaluation}. Since most of the sentence pairs obtained in this way fell into the first two categories, in order to balance the dataset, some of the examples were manually modified to convey different information. In this way, even negative examples often have high semantic overlap, making this problem difficult for machine learning models.

\subsection{Other datasets}
Below we briefly describe the other datasets used in our experiments:
\begin{itemize}[wide,labelwidth=0pt,labelindent=0pt]
\item \emph{Wroclaw Corpus of Consumer Reviews Sentiment (WCCRS)} \citep{Kocon2019} - A Polish sentiment analysis dataset containing consumer reviews of products and services, assigned to four classes: \emph{positive}, \emph{negative}, \emph{neutral} and \emph{ambiguous}. The corpus contains opinions relating to four domains: hotels, medical services, products, and education. Two of those domains, hotels and medical services, contain sentence-level annotations. We include them in our evaluation as \emph{Consumer Reviews - Hotels (CR-H)} and  \emph{Consumer Reviews - Medicine (CR-M)} tasks.
\item \emph{Sentences Involving Compositional Knowledge (SICK)} \citep{dadas-etal-2020-evaluation} - This corpus is a manually translated version of English SICK dataset \cite{marelli2014a}, containing 10,000 sentence pair with two types of annotations. Each pair contains a numerical score of semantic relatedness between sentences, and additionally a natural language inference (NLI) category label: \emph{entailment}, \emph{neutral} or \emph{contradiction}. We therefore include two evaluation tasks based on this dataset: classification \emph{(SICK-E)} and regression \emph{(SICK-R)}.
\item \emph{Compositional Distributional Semantics (CDS)} \citep{wroblewska-krasnowska-kieras-2017-polish} - A different corpus using the same annotation format as SICK dataset. Like the original, this corpus also contains 10,000 examples annotated with semantic relatedness score and NLI label. As with the previously described dataset, we also include two evaluation tasks in this case: \emph{CDS-E} and \emph{CDS-R}.
\item \emph{8TAGS} \citep{dadas-etal-2020-evaluation} - A collection of sentences relating to popular topics discussed on the Internet. It contains about 50,000 sentences annotated with 8 topic labels: \emph{film}, \emph{history}, \emph{food}, \emph{medicine}, \emph{motorization}, \emph{work}, \emph{sport} and \emph{technology}. A multi-class classification task \emph{(8TAGS)} based on this corpus was included in our experiments.
\end{itemize}

\subsection{Details of the experiments}
Neural sentence encoders developed as part of this study were trained on an automatically extracted collection of over 7 million sentence pairs. The dataset was constructed by running our paraphrase mining algorithm on English-Polish bilingual data from OpenSubtitles \citep{lison-tiedemann-2016-opensubtitles2016} and CCMatrix \citep{schwenk-etal-2021-ccmatrix}. We trained four neural architectures with different pooling layers. The first model was based on standard mean pooling, while the other three employed pooling using the LSTM layer with increasing cell memory size: 1024, 2048, and 4096. The weights of all sentence encoders were initialized with a pre-trained Polish RoBERTa base language model \citep{dadas2020pre} and the same set of hyperparameters was used for fine-tuning. We used a mini-batch size of 64. We employed a training scheduler with a linearly decreasing learning rate and a warmup phase for the first 10\% update steps. The peak learning rate was set to $2\mathrm{e}{-6}$. Each model was fine-tuned for three epochs, which took about 24-hours on a single Nvidia V100 GPU.

All datasets used in the evaluation have separate training, validation, and test parts. The results reported by us refer to the performance of the models on the test split of each dataset. As in the case of SentEval \citep{conneau-kiela-2018-senteval}, the training part is used for training a single layer classifier on top of the generated sentence embeddings, and the validation part is used for selecting optimal regularization term for this classifier.

In addition to the models described in this paper, we test several other neural sentence representations for Polish. We verify how the pre-trained neural language models perform on sentence-level tasks without additional fine-tuning on paraphrase data. For Polish, there are two high-quality models that were included in our study, Polish RoBERTa \citep{dadas2020pre} and HerBERT \citep{mroczkowski-etal-2021-herbert}, each of them available in base and large variants. We also test all available multilingual models from Sentence-Transformers library, fine-tuned using multilingual knowledge distillation method \citep{reimers-gurevych-2020-making}. Finally, we include the other commonly used neural sentence encoders: LASER \citep{10.1162/tacl_a_00288}, mUSE \citep{yang-etal-2020-multilingual} and LaBSE \citep{feng2020language}.

\subsection{Results}

\begin{table*}[htbp]
\caption{Evaluation of sentence representations on eight tasks for Polish. For classification tasks we use accuracy as the evaluation metric, for semantic relatedness tasks (\emph{CDSC-R} and \emph{SICK-R}) the Spearman's rank correlation coefficient is used.}
    \renewcommand{\arraystretch}{1.2}
    \begin{center}
    \begin{tabular}{c|c|c|cc|cc|cc|c}
        \toprule
        & & \makecell[c]{\textbf{Paraphrase}\\\textbf{identification}}  & \multicolumn{2}{c|}{\makecell[c]{\textbf{Sentiment}\\\textbf{analysis}}} & \multicolumn{2}{c|}{\makecell[c]{\textbf{Natural language}\\\textbf{inference}}} & \multicolumn{2}{c|}{\makecell[c]{\textbf{Semantic}\\\textbf{relatedness}}} & \makecell[c]{\textbf{Topic}\\\textbf{classification}} \\
        \hline
        \makecell[l]{\scriptsize\textbf{Model}} & \scriptsize\textbf{Avg.} & \scriptsize\textbf{PPC} & \scriptsize\textbf{CR-H} & \scriptsize\textbf{CR-M} &  \scriptsize\textbf{CDSC-E} & \scriptsize\textbf{SICK-E} & \scriptsize\textbf{CDSC-R} & \scriptsize\textbf{SICK-R} & \scriptsize\textbf{8TAGS} \\
        \hline
        \multicolumn{10}{l}{\textbf{Pre-trained language models + mean pooling}} \\
        \hline
        \makecell[l]{Polish RoBERTa (base)} & 75.13 & \textbf{70.60} & 85.14 & 79.21 & 81.40 & 71.89 & 79.53 & 62.26 & 71.00 \\
        \makecell[l]{HerBERT (base)} & 75.58 & 67.10 & 86.62 & 82.32 & 84.80 & 68.55 & 83.66 & 56.86 & 74.70 \\
        \makecell[l]{Polish RoBERTa (large)} & 76.64 & 65.40 & \textbf{87.90} & \textbf{83.56} & 82.60 & 68.59 & 83.89 & 61.84 & \textbf{79.30} \\
        \makecell[l]{HerBERT (large)} & \textbf{78.84} & 66.40 & 87.58 & 83.31 & \textbf{84.90} & \textbf{75.13} & \textbf{85.27} & \textbf{69.19} & 78.91 \\
        \hline
        \multicolumn{10}{l}{\textbf{Sentence encoders (Sentence-Transformers library)}} \\
        \hline
        \makecell[l]{paraphrase-multilingual-MiniLM-L12-v2} & 78.62 & 76.80 & 80.24 & 78.81 & 85.70 & 78.45 & 88.13 & 71.60 & 69.19 \\
        \makecell[l]{distilbert-multilingual-nli-stsb-quora-rank} & 78.80 & 80.00 & 80.44 & 74.02 & 86.70 & 79.66 & 87.23 & 73.61 & 68.73 \\
        \makecell[l]{distiluse-base-multilingual-cased-v2} & 78.85 & 73.80 & 79.92 & 75.95 & \textbf{87.90} & 78.68 & \textbf{90.54} & 73.12 & 70.86 \\
        \makecell[l]{xlm-r-bert-base-nli-stsb-mean-tokens} & 79.91 & 80.10 & 81.15 & 80.69 & 86.00 & 81.02 & 85.69 & 75.25 & 69.37 \\
        \makecell[l]{paraphrase-xlm-r-multilingual-v1} & 81.44 & \textbf{82.80} & 82.50 & 80.74 & 86.70 & \textbf{82.08} & 90.14 & \textbf{76.08} & 70.49\\
        \makecell[l]{paraphrase-multilingual-mpnet-base-v2} & \textbf{81.72} & 81.30 & \textbf{85.46} & \textbf{82.67} & 86.70 & 79.49 & 89.95 & 75.68 & \textbf{72.51}\\
        \hline
        \multicolumn{10}{l}{\textbf{Sentence encoders (other methods)}} \\
        \hline
        \makecell[l]{mUSE} & 78.79 & 72.70 & 79.02 & 74.22 & 86.30 & 81.96 & \textbf{90.23} & 76.61 & 69.26 \\
        \makecell[l]{LASER} & 80.07 & \textbf{80.10} & 81.21 & 78.12 & \textbf{87.90} & \textbf{82.21} & 89.30 & \textbf{76.65} & 65.07 \\
        \makecell[l]{LaBSE} & \textbf{81.20} & 77.10 & \textbf{86.10} & \textbf{80.64} & 86.70 & 81.63 & 90.02 & 76.04 & \textbf{71.36} \\
        \hline
        \multicolumn{10}{l}{\textbf{Our sentence encoder with different pooling layers}} \\
        \hline
        \makecell[l]{Mean pooling} & 81.53 & 79.40 & 85.78 & 80.94 & \textbf{87.90} & 80.35 & 88.57 & 76.56 & 72.78\\
        \makecell[l]{LSTM pooling (1024)} & 81.74 & 80.00 & 85.07 & 81.38 & 86.90 & 80.55 & 89.46 & 77.40 & \textbf{73.19}\\
        \makecell[l]{LSTM pooling (2048)} & 82.48 & \textbf{83.40} & 85.26 & \textbf{81.58} & 87.30 & 81.49 & 89.70 & 78.48 & 72.60\\
        \makecell[l]{LSTM pooling (4096)} & \textbf{82.52} & 82.50 & \textbf{86.49} & 81.04 & 86.90 & \textbf{82.39} & \textbf{90.06} & \textbf{78.99} & 71.75\\
        \bottomrule
    \end{tabular}
    \end{center}
    \label{tab_results}
\end{table*}

The results of our experiments are shown in Table \ref{tab_results}. The table is divided into four sections, each corresponding to specific groups of sentence encoding methods. The first group includes pre-trained neural language models based on the Transformer architecture. In this part of the evaluation, we utilized the original model weights without fine-tuning them for encoding sentences. In this case, sentence representation is constructed by aggregating individual token vectors generated by the last layer of the network using mean pooling operation. As we can see, these representations perform well on conventional classification tasks such as sentiment analysis or topic classification but perform considerably worse on tasks that require measuring semantic relationship between sentences. On the \emph{SICK}, \emph{CDSC} and \emph{PPC} datasets, the results are up to 13\% lower compared to the other evaluated solutions.

The second group consists of sentence encoders from the Sentence-Transformers library trained using multilingual knowledge distillation method \citep{reimers-gurevych-2020-making}. These models offer varying performance on Polish language tasks. However, two of them stand out: \emph{paraphrase-xlm-r-multilingual-v1} and \emph{paraphrase-multilingual-mpnet-base-v2}. Both achieved an average score on all tasks above 81\%. The former was particularly good on the semantic retrieval tasks while scoring lower on typical classification tasks. The latter, on the other hand, obtained balanced scores on all types of tasks.

Three popular neural sentence embedding methods published by Google and Facebook are included in the third group. Each of them performs well on semantic relatedness and natural language inference problems, while results on the other tasks often fall short of the performance offered by competing approaches. Of the three architectures compared, LaBSE performed best by achieving an average score above 81\%. Reasonable performance is also offered by the LASER model with an average score of around 80\%. However, in both cases their results were lower than those of the best models available in the Sentence-Transformers library.

The last section presents the results of our sentence encoders applying different pooling methods. The model using standard mean pooling achieved an average score of 81.53\%, which is the second-best result in comparison to the previously discussed methods. Models incorporating an additional layer of LSTM-based pooling further improved this result. We can see that increasing the dimension of the sentence embedding has a positive effect on model performance. The difference is especially noticeable between LSTM layers with a hidden size of 1024 and 2048, while increasing the dimension further brings smaller improvements. The two largest models achieved an average score over all tasks of about 82.5\%. It is also worth noting that all our models offer balanced performance across different tasks. On conventional classification problems, only pre-trained language models perform better. Whereas on paraphrase identification, natural language inference, and semantic relatedness, our encoders perform just as well or better than the best multilingual sentence encoders.

\section{Conclusion}
In this paper, we have proposed a method for training neural sentence encoders without manually labeled data. Our approach involves automatic extraction of paraphrase pairs from sentence-aligned bilingual text corpora. Such datasets are readily available for many languages, so our technique is particularly suitable for training models in low-resource languages for which there is little or no annotated data. Using the extracted sentence pairs, we fine-tune a language model based on the Transformer architecture with an additional recurrent pooling layer responsible for generating sentence embeddings. The method allows us to train an effective language-specific sentence encoder in a short time on a single GPU, outperforming state-of-the-art multilingual models whose training required significantly more computational resources.

We validated our approach on eight tasks for the Polish language. For evaluation purposes, we also developed a new dataset that includes 7000 manually annotated sentence pairs, the Polish Paraphrase Corpus (PPC). Our sentence embeddings have shown high quality on a variety of linguistic problems, performing well both on semantic retrieval tasks and on sentence-level classification tasks.

\footnotesize\bibliography{references}

\begin{thebibliography}{38}
\providecommand{\natexlab}[1]{#1}
\providecommand{\url}[1]{#1}
\csname url@samestyle\endcsname
\providecommand{\newblock}{\relax}
\providecommand{\bibinfo}[2]{#2}
\providecommand{\BIBentrySTDinterwordspacing}{\spaceskip=0pt\relax}
\providecommand{\BIBentryALTinterwordstretchfactor}{4}
\providecommand{\BIBentryALTinterwordspacing}{\spaceskip=\fontdimen2\font plus
\BIBentryALTinterwordstretchfactor\fontdimen3\font minus
  \fontdimen4\font\relax}
\providecommand{\BIBforeignlanguage}[2]{{%
\expandafter\ifx\csname l@#1\endcsname\relax
\typeout{** WARNING: IEEEtranN.bst: No hyphenation pattern has been}%
\typeout{** loaded for the language `#1'. Using the pattern for}%
\typeout{** the default language instead.}%
\else
\language=\csname l@#1\endcsname
\fi
#2}}
\providecommand{\BIBdecl}{\relax}
\BIBdecl

\bibitem[Bowman et~al.(2015)Bowman, Angeli, Potts, and
  Manning]{bowman2015large}
S.~Bowman, G.~Angeli, C.~Potts, and C.~D. Manning, ``A large annotated corpus
  for learning natural language inference,'' in \emph{Proceedings of the 2015
  Conference on Empirical Methods in Natural Language Processing}, 2015, pp.
  632--642.

\bibitem[Williams et~al.(2018)Williams, Nangia, and Bowman]{williams2018broad}
A.~Williams, N.~Nangia, and S.~Bowman, ``A broad-coverage challenge corpus for
  sentence understanding through inference,'' in \emph{Proceedings of the 2018
  Conference of the North American Chapter of the Association for Computational
  Linguistics: Human Language Technologies, Volume 1 (Long Papers)}, 2018, pp.
  1112--1122.

\bibitem[Tiedemann(2012)]{tiedemann2012parallel}
J.~Tiedemann, ``Parallel data, tools and interfaces in opus,'' in
  \emph{Proceedings of the Eighth International Conference on Language
  Resources and Evaluation (LREC'12)}, 2012, pp. 2214--2218.

\bibitem[Vaswani et~al.(2017)Vaswani, Shazeer, Parmar, Uszkoreit, Jones, Gomez,
  Kaiser, and Polosukhin]{vaswani2017attention}
A.~Vaswani, N.~Shazeer, N.~Parmar, J.~Uszkoreit, L.~Jones, A.~N. Gomez,
  {\L}.~Kaiser, and I.~Polosukhin, ``Attention is all you need,'' in
  \emph{Advances in neural information processing systems}, 2017, pp.
  5998--6008.

\bibitem[Hochreiter and Schmidhuber(1997)]{hochreiter1997long}
S.~Hochreiter and J.~Schmidhuber, ``Long short-term memory,'' \emph{Neural
  computation}, vol.~9, no.~8, pp. 1735--1780, 1997.

\bibitem[Mikolov et~al.(2013{\natexlab{a}})Mikolov, Chen, Corrado, and
  Dean]{mikolov2013efficient}
T.~Mikolov, K.~Chen, G.~S. Corrado, and J.~Dean, ``Efficient estimation of word
  representations in vector space,'' \emph{International Conference on Learning
  Representations}, 2013.

\bibitem[Mikolov et~al.(2013{\natexlab{b}})Mikolov, Sutskever, Chen, Corrado,
  and Dean]{mikolov2013distributed}
T.~Mikolov, I.~Sutskever, K.~Chen, G.~S. Corrado, and J.~Dean, ``Distributed
  representations of words and phrases and their compositionality,''
  \emph{Neural Information Processing Systems}, pp. 3111--3119, 2013.

\bibitem[Pennington et~al.(2014)Pennington, Socher, and
  Manning]{pennington2014glove}
J.~Pennington, R.~Socher, and C.~D. Manning, ``Glove: Global vectors for word
  representation,'' in \emph{Proceedings of the 2014 Conference on Empirical
  Methods in Natural Language Processing (EMNLP)}.\hskip 1em plus 0.5em minus
  0.4em\relax Empirical Methods in Natural Language Processing, 2014, pp.
  1532--1543.

\bibitem[Bojanowski et~al.(2017)Bojanowski, Grave, Joulin, and
  Mikolov]{bojanowski2017enriching}
P.~Bojanowski, E.~Grave, A.~Joulin, and T.~Mikolov, ``Enriching word vectors
  with subword information,'' \emph{Transactions of the Association for
  Computational Linguistics}, pp. 135--146, 2017.

\bibitem[Arora et~al.(2017)Arora, Liang, and Ma]{arora2017a}
S.~Arora, Y.~Liang, and T.~Ma, ``A simple but tough-to-beat baseline for
  sentence embeddings,'' in \emph{ICLR 2017 : International Conference on
  Learning Representations 2017}.\hskip 1em plus 0.5em minus 0.4em\relax
  International Conference on Learning Representations, 2017.

\bibitem[Shen et~al.(2018)Shen, Wang, Wang, Min, Su, Zhang, Li, Henao, and
  Carin]{shen2018baseline}
D.~Shen, G.~Wang, W.~Wang, M.~R. Min, Q.~Su, Y.~Zhang, C.~Li, R.~Henao, and
  L.~Carin, ``Baseline needs more love: On simple word-embedding-based models
  and associated pooling mechanisms,'' \emph{Meeting of the Association for
  Computational Linguistics}, pp. 440--450, 2018.

\bibitem[Le and Mikolov(2014)]{le2014distributed}
Q.~Le and T.~Mikolov, ``Distributed representations of sentences and
  documents,'' in \emph{International conference on machine learning}.\hskip
  1em plus 0.5em minus 0.4em\relax PMLR, 2014, pp. 1188--1196.

\bibitem[Kiros et~al.(2015)Kiros, Zhu, Salakhutdinov, Zemel, Torralba, Urtasun,
  and Fidler]{kiros2015skipthought}
R.~Kiros, Y.~Zhu, R.~R. Salakhutdinov, R.~S. Zemel, A.~Torralba, R.~Urtasun,
  and S.~Fidler, ``Skip-thought vectors,'' \emph{Neural Information Processing
  Systems}, pp. 3294--3302, 2015.

\bibitem[Gan et~al.(2017)Gan, Pu, Henao, Li, He, and Carin]{gan2017learning}
Z.~Gan, Y.~Pu, R.~Henao, C.~Li, X.~He, and L.~Carin, ``Learning generic
  sentence representations using convolutional neural networks,'' in
  \emph{Proceedings of the 2017 Conference on Empirical Methods in Natural
  Language Processing}.\hskip 1em plus 0.5em minus 0.4em\relax Empirical
  Methods in Natural Language Processing, 2017, pp. 2390--2400.

\bibitem[Logeswaran and Lee(2018)]{logeswaran2018an}
L.~Logeswaran and H.~Lee, ``An efficient framework for learning sentence
  representations,'' \emph{International Conference on Learning
  Representations}, 2018.

\bibitem[Wang et~al.(2021)Wang, Reimers, and
  Gurevych]{wang-etal-2021-tsdae-using}
K.~Wang, N.~Reimers, and I.~Gurevych, ``{TSDAE}: Using transformer-based
  sequential denoising auto-encoderfor unsupervised sentence embedding
  learning,'' in \emph{Findings of the Association for Computational
  Linguistics: EMNLP 2021}.\hskip 1em plus 0.5em minus 0.4em\relax Punta Cana,
  Dominican Republic: Association for Computational Linguistics, Nov. 2021, pp.
  671--688.

\bibitem[Gao et~al.(2021)Gao, Yao, and Chen]{gao-etal-2021-simcse}
T.~Gao, X.~Yao, and D.~Chen, ``{S}im{CSE}: Simple contrastive learning of
  sentence embeddings,'' in \emph{Proceedings of the 2021 Conference on
  Empirical Methods in Natural Language Processing}.\hskip 1em plus 0.5em minus
  0.4em\relax Online and Punta Cana, Dominican Republic: Association for
  Computational Linguistics, Nov. 2021, pp. 6894--6910.

\bibitem[Carlsson et~al.(2021)Carlsson, Gyllensten, Gogoulou, Hellqvist, and
  Sahlgren]{carlsson2021semantic}
F.~Carlsson, A.~C. Gyllensten, E.~Gogoulou, E.~Y. Hellqvist, and M.~Sahlgren,
  ``Semantic re-tuning with contrastive tension,'' in \emph{International
  Conference on Learning Representations}, 2021.

\bibitem[Conneau et~al.(2017)Conneau, Kiela, Schwenk, Barrault, and
  Bordes]{conneau2017supervised}
A.~Conneau, D.~Kiela, H.~Schwenk, L.~Barrault, and A.~Bordes, ``Supervised
  learning of universal sentence representations from natural language
  inference data,'' in \emph{Proceedings of the 2017 Conference on Empirical
  Methods in Natural Language Processing}.\hskip 1em plus 0.5em minus
  0.4em\relax Empirical Methods in Natural Language Processing, 2017, pp.
  670--680.

\bibitem[Cer et~al.(2018)Cer, Yang, Kong, Hua, Limtiaco, John, Constant,
  Guajardo-C{\'{e}}spedes, Yuan, Tar, Sung, Strope, and
  Kurzweil]{cer2018universal}
D.~Cer, Y.~Yang, S.-y. Kong, N.~Hua, N.~L.~U. Limtiaco, R.~S. John,
  N.~Constant, M.~Guajardo-C{\'{e}}spedes, S.~Yuan, C.~Tar, Y.-h. Sung,
  B.~Strope, and R.~Kurzweil, ``Universal sentence encoder,'' \emph{arXiv
  preprint arXiv:1803.11175}, 2018.

\bibitem[Reimers et~al.(2019)Reimers, Gurevych, Reimers, Gurevych, Thakur,
  Reimers, Daxenberger, Gurevych, Reimers, Gurevych,
  et~al.]{reimers2019sentence}
N.~Reimers, I.~Gurevych, N.~Reimers, I.~Gurevych, N.~Thakur, N.~Reimers,
  J.~Daxenberger, I.~Gurevych, N.~Reimers, I.~Gurevych \emph{et~al.},
  ``Sentence-bert: Sentence embeddings using siamese bert-networks,'' in
  \emph{Proceedings of the 2019 Conference on Empirical Methods in Natural
  Language Processing}.\hskip 1em plus 0.5em minus 0.4em\relax Association for
  Computational Linguistics, 2019.

\bibitem[Artetxe and Schwenk(2019)]{10.1162/tacl_a_00288}
M.~Artetxe and H.~Schwenk, ``{Massively Multilingual Sentence Embeddings for
  Zero-Shot Cross-Lingual Transfer and Beyond},'' \emph{Transactions of the
  Association for Computational Linguistics}, vol.~7, pp. 597--610, 09 2019.

\bibitem[Yang et~al.(2020)Yang, Cer, Ahmad, Guo, Law, Constant,
  Hernandez~Abrego, Yuan, Tar, Sung, Strope, and
  Kurzweil]{yang-etal-2020-multilingual}
Y.~Yang, D.~Cer, A.~Ahmad, M.~Guo, J.~Law, N.~Constant, G.~Hernandez~Abrego,
  S.~Yuan, C.~Tar, Y.-h. Sung, B.~Strope, and R.~Kurzweil, ``Multilingual
  universal sentence encoder for semantic retrieval,'' in \emph{Proceedings of
  the 58th Annual Meeting of the Association for Computational Linguistics:
  System Demonstrations}.\hskip 1em plus 0.5em minus 0.4em\relax Online:
  Association for Computational Linguistics, Jul. 2020, pp. 87--94.

\bibitem[Feng et~al.(2020)Feng, Yang, Cer, Arivazhagan, and
  Wang]{feng2020language}
F.~Feng, Y.~Yang, D.~Cer, N.~Arivazhagan, and W.~Wang, ``Language-agnostic bert
  sentence embedding,'' \emph{arXiv preprint arXiv:2007.01852}, 2020.

\bibitem[Devlin et~al.(2019)Devlin, Chang, Lee, and
  Toutanova]{devlin-etal-2019-bert}
J.~Devlin, M.-W. Chang, K.~Lee, and K.~Toutanova, ``{BERT}: Pre-training of
  deep bidirectional transformers for language understanding,'' in
  \emph{Proceedings of the 2019 Conference of the North {A}merican Chapter of
  the Association for Computational Linguistics: Human Language Technologies,
  Volume 1 (Long and Short Papers)}.\hskip 1em plus 0.5em minus 0.4em\relax
  Minneapolis, Minnesota: Association for Computational Linguistics, Jun. 2019,
  pp. 4171--4186.

\bibitem[Reimers and Gurevych(2020)]{reimers-gurevych-2020-making}
N.~Reimers and I.~Gurevych, ``Making monolingual sentence embeddings
  multilingual using knowledge distillation,'' in \emph{Proceedings of the 2020
  Conference on Empirical Methods in Natural Language Processing
  (EMNLP)}.\hskip 1em plus 0.5em minus 0.4em\relax Online: Association for
  Computational Linguistics, Nov. 2020, pp. 4512--4525.

\bibitem[Lison and Tiedemann(2016)]{lison-tiedemann-2016-opensubtitles2016}
P.~Lison and J.~Tiedemann, ``{O}pen{S}ubtitles2016: Extracting large parallel
  corpora from movie and {TV} subtitles,'' in \emph{Proceedings of the Tenth
  International Conference on Language Resources and Evaluation
  ({LREC}'16)}.\hskip 1em plus 0.5em minus 0.4em\relax Portoro{\v{z}},
  Slovenia: European Language Resources Association (ELRA), May 2016, pp.
  923--929.

\bibitem[Schwenk et~al.(2021)Schwenk, Wenzek, Edunov, Grave, Joulin, and
  Fan]{schwenk-etal-2021-ccmatrix}
H.~Schwenk, G.~Wenzek, S.~Edunov, E.~Grave, A.~Joulin, and A.~Fan,
  ``{CCM}atrix: Mining billions of high-quality parallel sentences on the
  web,'' in \emph{Proceedings of the 59th Annual Meeting of the Association for
  Computational Linguistics and the 11th International Joint Conference on
  Natural Language Processing (Volume 1: Long Papers)}.\hskip 1em plus 0.5em
  minus 0.4em\relax Online: Association for Computational Linguistics, Aug.
  2021, pp. 6490--6500.

\bibitem[Liu et~al.(2019)Liu, Ott, Goyal, Du, Joshi, Chen, Levy, Lewis,
  Zettlemoyer, and Stoyanov]{liu2019roberta}
Y.~Liu, M.~Ott, N.~Goyal, J.~Du, M.~Joshi, D.~Chen, O.~Levy, M.~Lewis,
  L.~Zettlemoyer, and V.~Stoyanov, ``Roberta: A robustly optimized bert
  pretraining approach,'' \emph{arXiv preprint arXiv:1907.11692}, 2019.

\bibitem[Loshchilov and Hutter(2019)]{loshchilov2018decoupled}
I.~Loshchilov and F.~Hutter, ``Decoupled weight decay regularization,'' in
  \emph{International Conference on Learning Representations}, 2019.

\bibitem[Henderson et~al.(2017)Henderson, Al-Rfou, Strope, Sung, Luk{\'a}cs,
  Guo, Kumar, Miklos, and Kurzweil]{Henderson2017EfficientNL}
M.~Henderson, R.~Al-Rfou, B.~Strope, Y.-H. Sung, L.~Luk{\'a}cs, R.~Guo,
  S.~Kumar, B.~Miklos, and R.~Kurzweil, ``Efficient natural language response
  suggestion for smart reply,'' \emph{ArXiv}, vol. abs/1705.00652, 2017.

\bibitem[Conneau and Kiela(2018)]{conneau-kiela-2018-senteval}
A.~Conneau and D.~Kiela, ``{S}ent{E}val: An evaluation toolkit for universal
  sentence representations,'' in \emph{Proceedings of the Eleventh
  International Conference on Language Resources and Evaluation ({LREC}
  2018)}.\hskip 1em plus 0.5em minus 0.4em\relax Miyazaki, Japan: European
  Language Resources Association (ELRA), May 2018.

\bibitem[Dadas et~al.(2020{\natexlab{a}})Dadas, Pere{\l}kiewicz, and
  Po{\'s}wiata]{dadas-etal-2020-evaluation}
S.~Dadas, M.~Pere{\l}kiewicz, and R.~Po{\'s}wiata,
  ``\BIBforeignlanguage{English}{Evaluation of sentence representations in
  {P}olish},'' in \emph{\BIBforeignlanguage{English}{Proceedings of the 12th
  Language Resources and Evaluation Conference}}.\hskip 1em plus 0.5em minus
  0.4em\relax Marseille, France: European Language Resources Association, May
  2020, pp. 1674--1680.

\bibitem[Koco{\'n} et~al.(2019)Koco{\'n}, Zaśko-Zielińska, and
  Miłkowski]{Kocon2019}
J.~Koco{\'n}, M.~Zaśko-Zielińska, and P.~Miłkowski, ``{Multi-level analysis
  and recognition of the text sentiment on the example of consumer opinions},''
  in \emph{Proceedings of the International Conference Recent Advances in
  Natural Language Processing, RANLP 2019}, 2019.

\bibitem[Marelli et~al.(2014)Marelli, Menini, Baroni, Bentivogli, bernardi, and
  Zamparelli]{marelli2014a}
M.~Marelli, S.~Menini, M.~Baroni, L.~Bentivogli, R.~bernardi, and
  R.~Zamparelli, ``A sick cure for the evaluation of compositional
  distributional semantic models,'' in \emph{Proceedings of the Ninth
  International Conference on Language Resources and Evaluation
  (LREC'14)}.\hskip 1em plus 0.5em minus 0.4em\relax Language Resources and
  Evaluation, 2014, pp. 216--223.

\bibitem[Wr{\'o}blewska and
  Krasnowska-Kiera{\'s}(2017)]{wroblewska-krasnowska-kieras-2017-polish}
A.~Wr{\'o}blewska and K.~Krasnowska-Kiera{\'s}, ``{P}olish evaluation dataset
  for compositional distributional semantics models,'' in \emph{Proceedings of
  the 55th Annual Meeting of the Association for Computational Linguistics
  (Volume 1: Long Papers)}.\hskip 1em plus 0.5em minus 0.4em\relax Vancouver,
  Canada: Association for Computational Linguistics, Jul. 2017, pp. 784--792.

\bibitem[Dadas et~al.(2020{\natexlab{b}})Dadas, Pere{\l}kiewicz, and
  Po{\'s}wiata]{dadas2020pre}
S.~Dadas, M.~Pere{\l}kiewicz, and R.~Po{\'s}wiata, ``Pre-training polish
  transformer-based language models at scale,'' in \emph{International
  Conference on Artificial Intelligence and Soft Computing}.\hskip 1em plus
  0.5em minus 0.4em\relax Springer, 2020, pp. 301--314.

\bibitem[Mroczkowski et~al.(2021)Mroczkowski, Rybak, Wr{\'o}blewska, and
  Gawlik]{mroczkowski-etal-2021-herbert}
R.~Mroczkowski, P.~Rybak, A.~Wr{\'o}blewska, and I.~Gawlik, ``{H}er{BERT}:
  Efficiently pretrained transformer-based language model for {P}olish,'' in
  \emph{Proceedings of the 8th Workshop on Balto-Slavic Natural Language
  Processing}.\hskip 1em plus 0.5em minus 0.4em\relax Kiyv, Ukraine:
  Association for Computational Linguistics, Apr. 2021, pp. 1--10.

\end{thebibliography}

\end{document}